\documentclass[letterpaper]{article} 
\usepackage{aaai24}  
\usepackage{times}  
\usepackage{helvet}  
\usepackage{courier}  
\usepackage[hyphens]{url}  
\usepackage{graphicx} 
\usepackage{natbib}  
\usepackage{caption} 
\usepackage[vlined,linesnumbered]{algorithm2e}

\usepackage{times}
\usepackage{epsfig}
\usepackage{graphicx}
\usepackage{amsmath}
\usepackage{amssymb}
\usepackage{subfig}
\usepackage{booktabs}
\usepackage{multirow}
\usepackage{makecell}
\usepackage{color}
\usepackage[table]{xcolor}
\usepackage{newfloat}
\usepackage{listings}
\DeclareCaptionStyle{ruled}{labelfont=normalfont,labelsep=colon,strut=off} 
\newcommand{\ie}{\textit{i.e.}}
\title{DiffusionTrack: Diffusion Model For Multi-Object Tracking}
\author {
    Run Luo\textsuperscript{\rm 123},
    Zikai Song\textsuperscript{\rm 3\footnote{co-corresponding author}},
    Lintao Ma\textsuperscript{\rm 3},
    Jinlin Wei\textsuperscript{\rm 34},
    Wei Yang\textsuperscript{\rm 3}, Min Yang\textsuperscript{\rm 12\footnotemark[1]} \\
}
\affiliations {
    \textsuperscript{\rm 1}Shenzen Institute of Advanced Technology, Chinese Academy of Sciences\\
    \textsuperscript{\rm 2}University of Chinese Academy of Sciences\\
    \textsuperscript{\rm 3}Huazhong University of Science and Technology\\
    \textsuperscript{\rm 4}University of California, Santa Barbara\\
    
    \{r.luo,min.yang\}@siat.ac.cn, \{skyesong, mltdml,weiyangcs\}@hust.edu.cn, jinlinwei@ucsb.edu \\
}

\usepackage{bibentry}

\begin{document}

\maketitle

\begin{abstract}
Multi-object tracking (MOT) is a challenging vision task that aims to detect individual objects within a single frame and associate them across multiple frames. Recent MOT approaches can be categorized into two-stage tracking-by-detection (TBD) methods and one-stage joint detection and tracking (JDT) methods. Despite the success of these approaches, they also suffer from common problems, such as harmful global or local inconsistency, poor trade-off between robustness and model complexity, and lack of flexibility in different scenes within the same video. In this paper we propose a simple but robust framework that formulates object detection and association jointly as a consistent denoising diffusion process from paired noise boxes to paired ground-truth boxes. This novel progressive denoising diffusion strategy substantially augments the tracker's effectiveness, enabling it to discriminate between various objects. During the training stage, paired object boxes diffuse from paired ground-truth boxes to random distribution, and the model learns detection and tracking simultaneously by reversing this noising process. In inference, the model refines a set of paired randomly generated boxes to the detection and tracking results in a flexible one-step or multi-step denoising diffusion process. Extensive experiments on three widely used MOT benchmarks, including MOT17, MOT20, and DanceTrack, demonstrate that our approach achieves competitive performance compared to the current state-of-the-art methods. Code is available at \textbf{\url{https://github.com/RainBowLuoCS/DiffusionTrack}}.
\end{abstract}

\section{Introduction}

Multi-object Tracking is one of the fundamental vision tasks with applications ranging from human-computer interaction, surveillance, autonomous driving, etc. It aims at detecting the bounding box of the object and associating the same object across consecutive frames in a video sequence. 
Recent MOT approaches can be categorized into two-stage tracking-by-detection (TBD) methods and one-stage joint detection and tracking (JDT) methods. 
TBD methods detect the bounding boxes of the objects within a single frame using a detector and associate the same object cross frames by employing supplementary trackers. These trackers encompass a spectrum of techniques, such as motion-based trackers~\cite{SORT,OCSORT,ByteTrack,BoTSORT,p3aformer,DeepSORT,FairMOT,SparseTrack} that employ the Kalman filter framework~\cite{KalmanFilter}. In addition, certain TBD approaches establish object associations through the utilization of Re-identification (Re-ID) techniques~\cite{chen2018real,bergmann2019tracking}, and others that rely on graph-based trackers~\cite{GMTracker,TrackMPNN,li2020graph} that model the association process as minimization of a cost flow problem.

\begin{figure}[t]
  \centering \includegraphics[width=0.92\columnwidth,keepaspectratio]{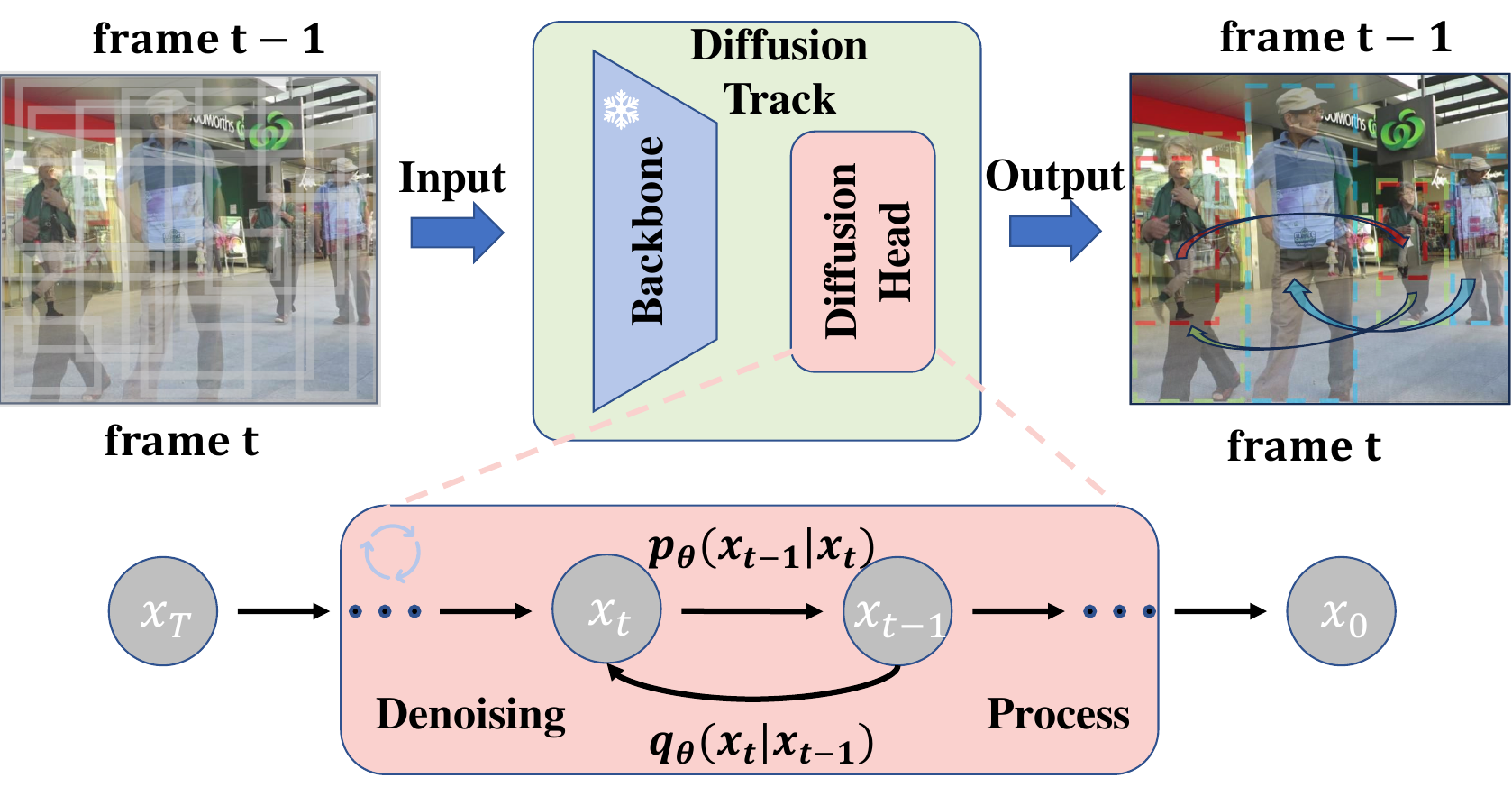}
  \caption{DiffusionTrack formulates object association as a denoising diffusion process from paired noise boxes to paired object boxes within two adjacent frames $t-1$ and $t$. The diffusion head receives the two-frame image information extracted by the frozen backbone and then iteratively denoises the paired noise boxes to obtain the final paired object boxes.}
  \label{fig1}
\end{figure}

JDT approaches try to combine the tracking and detection process in a unified manner. This paradigm consists of three mainstream strategies: query-based trackers \cite{TransTrack, TrackFormer, MOTR, MeMOT, TransT} that adopt unique query implicitly by forcing each query to track the same object, offset-based trackers \cite{Tracktor, PermaTrack, TransCenter, CenterTrack} utilizing the motion feature to predict motion offset, and trajectory-based trackers \cite{pang2020tubetk,zhou2022global} that tackle severe object occlusions via spatial-temporal information. However, most of TBD and JDT approaches suffer from the following common drawbacks: 
(1) Harmful global or local inconsistency plagues both methods. In TBD approaches, 
 the segmentation of detection and tracking tasks into distinct training processes engenders global inconsistencies that curtail overall performance. 
Although JDT approaches aim to bridge the gap between detection and tracking, they still treat them as disparate tasks through various branches or modules, not fully resolving the inconsistency;
(2) A suboptimal balance between robustness and model complexity is evident in both approaches. While the simple structure of TBD methods suffers from poor performance when faced with detection perturbation, the complex design of JDT approaches ensures stability and robustness but compromises detection accuracy compared to TBD methods;
(3) Both approaches also exhibit inflexibility across different scenes within the same video. Conventional methods process videos under uniform settings, hindering the adaptive application of strategies for varying scenes and consequently limiting their efficacy.

Recently, diffusion models have not only excelled in various generative tasks but also demonstrated potential in confronting complex discriminative computer vision challenges \cite{chen2022diffusiondet,gu2022diffusioninst}. This paper introduces DiffusionTrack, inspired by the progress in diffusion models, and constructs a novel consistent noise-to-tracking paradigm. DiffusionTrack directly formulates object associations from a set of paired random boxes within two adjacent frames, as illustrated in Figure \ref{fig1}. The motivation is to meticulously refine the coordinates of these paired boxes so that they accurately cover the same targeted objects across two consecutive frames, thereby implicitly performing detection and tracking within a uniform model pipeline. This innovative coarse-to-fine paradigm is believed to compel the model to learn to accurately distinguish objects from one another, ultimately leading to enhanced performance. DiffusionTrack addresses the multi-object tracking task by treating data association as a generative endeavor within the space of paired bounding boxes over two successive frames. Extensive experiments on 3 challenging datasets including MOT17~\cite{MOT16}, MOT20~\cite{MOT20} and DanceTrack~\cite{DanceTrack}, exhibit the state-of-the-art performance among the JDT multi-object trackers, which is also compared with TBD approaches. 

In summary, our main contributions include: 
\begin{enumerate}
\item We propose DiffusionTrack, which is the first work to employ the diffusion model for multi-object tracking by formulating it as a generative noise-to-tracking diffusion process.

\item Experimental results show that our noise-to-tracking paradigm has several appealing properties, such as decoupling training and evaluation stage for dynamic boxes and progressive refinement, promising consistency model structure for two tasks, and strong robustness to detection perturbation results.

\end{enumerate}

\section{Related Work}

Existing MOT algorithms can be divided into two categories according to the paradigm of handling the detection and association, \ie, the two-stage TBD methods and the one-stage JDT methods. 

\textbf{Two-stage TBD methods} is a common practice in the MOT field, where object detection and data association are treated as separate modules. The object detection module uses an existing detector~\cite{FasterRCNN, CenterNet, YOLOX}, and the data association module can be further divided into motion-based methods\cite{SORT, DeepSORT, ByteTrack, BoTSORT, OCSORT} and graph-based \cite{zhang2008global,jiang2019graph,braso2020learning,li2020graph, GMTracker} methods.
\textit{Motion-based methods} integrate detections through a distingct motion tracker across consecutive frames, employing various techniques.
SORT \cite{SORT} initialed the use of the Kalman filter \cite{KalmanFilter} for object tracking, associating each bounding box with the highest overlap through the Hungarian algorithm \cite{hungarian}. DeepSORT \cite{DeepSORT} enhanced this by incorporating both motion and deep appearance features, while StrongSORT~\cite{StrongSORT} further integrated lightweight, appearance-free algorithms for detection and association. ByteTrack~\cite{ByteTrack} addressed fragmented trajectories and missing detections by utilizing low-confidence detection similarities.
P3AFormer~\cite{p3aformer} combined pixel-wise distribution architecture with Kalman filter to refine object association, and OC-SORT~\cite{OCSORT} amended the linear motion assumption within the Klaman Filter for superior adaptability to occlusion and non-linear motion. 
\textit{Graph-based methods}, including Graph Neural Networks (GNN)~\cite{GNN} and Graph Convolutional Networks (GCN)~\cite{GCN}, have been widely explored in MOT, with vertices representing detection bounding boxes or tracklets and edges across frames denoting similarities. This setup allows the association challenge to be cast as a min-cost flow problem. MPNTrack ~\cite{braso2020learning} introduced a message-passing network to capture information between vertices across frames, GNMOT ~\cite{li2020graph} constructed dual graph networks to model appearance and motion features, and GMTracker ~\cite{GMTracker} emphasized both inter-frame matching and intra-frame context. 

\begin{figure*}
  \centering \includegraphics[width=0.95\textwidth,keepaspectratio,page=1]{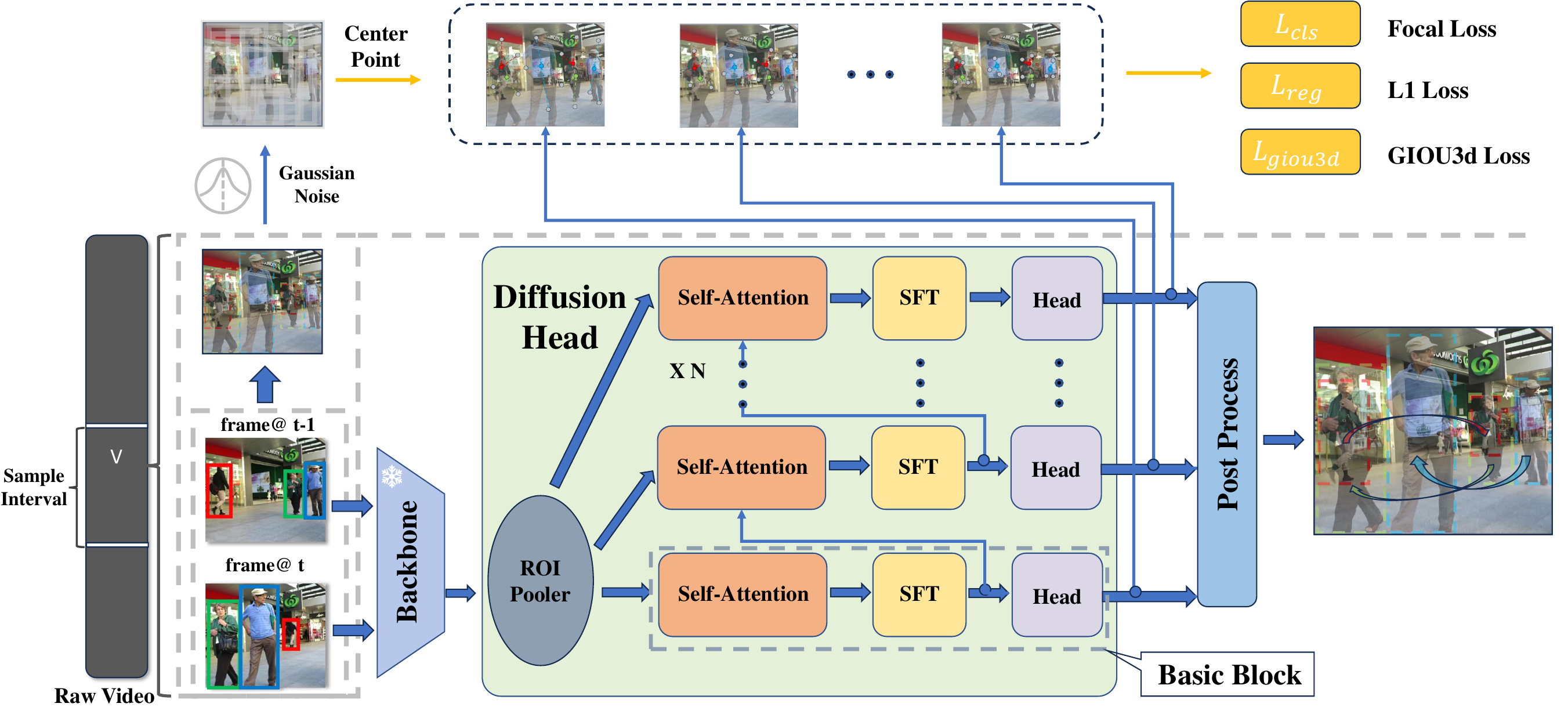}
  \caption{The architecture of DiffusionTrack. Given the images and corresponding ground-truth in the frame \textit{t} and frame \textit{t-1}, we extract features from two adjacent frames through the frozen backbone, then the diffusion head takes paired noise boxes as input and predicts category classification, box coordinates and association score of the same object in two adjacent frames. During training, the noise boxes are constructed by adding Gaussian noise to paired ground-truth boxes of the same object. In inference, the noise boxes are constructed by adding Gaussian noise to the padded prior object boxes in the previous frame.}
  \label{fig2}
\end{figure*}

\textbf{One-stage JDT methods.}
In recent years, there have been several explorations into the one-stage paradigm, which combines object detection and data association into a single pipeline.
\textit{Query-based methods}, a burgeoning trend, utilize DETR \cite{DETR, DeformableDETR} extensions for MOT by representing each object as a query regressed across various frames. Techniques such as TrackFormer \cite{TrackFormer} and MOTR \cite{MOTR} perform simultaneous object detection and association using concatenated object and track queries. TransTrack \cite{TransTrack} employs cyclical feature passing to aggregate embeddings, while MeMOT \cite{MeMOT} encodes historical observations to preserve extensive spatio-temporal memory.
\textit{Offset-based methods}, in contrast, bypass inter-frame association and instead focus on regressing past object locations to new positions. This approach includes Tracktor++ \cite{MeMOT} for temporal realignment of bounding boxes, CenterTrack \cite{CenterTrack} for object localization and offset prediction, and PermaTrack \cite{PermaTrack}, which fuses historical memory to reason target location and occlusion. TransCenter \cite{TransCenter} further advances this category by adopting dense representations with image-specific detection queries and tracking.
\textit{Trajectory-based methods} extract spatial-temporal information from historical tracklets to associate objects. GTR \cite{zhou2022global} groups detections from consecutive frames into trajectories using trajectory queries, and TubeTK \cite{pang2020tubetk} 
 extends bounding-boxes to video-based bounding-tubes for prediction. Both efficiently handle occlusion issues by utilizing long-term tracklet information.

\textbf{Diffusion model.} As a class of deep generative models, diffusion models \cite{ho2020denoising,song2019generative,song2020score} start from the sample in random distribution and recover the data sample via a gradual denoising process. 

However, their potential for visual understanding tasks has yet to be fully explored. Recently, DiffusionDet \cite{chen2022diffusiondet} and DiffusionInst \cite{gu2022diffusioninst} have successfully applied diffusion models to object detection and instance segmentation as noise-to-box and noise-to-filter tasks, respectively. Inspired by their successful application of the diffusion model, we proposed DiffusionTrack, which further broadens the application of the diffusion model by formalizing MOT as a denoising process. To the best of our knowledge, this is the first work that adopts a diffusion model for the MOT task.

\begin{figure*}
  \centering \includegraphics[width=0.92\textwidth,keepaspectratio,page=1]{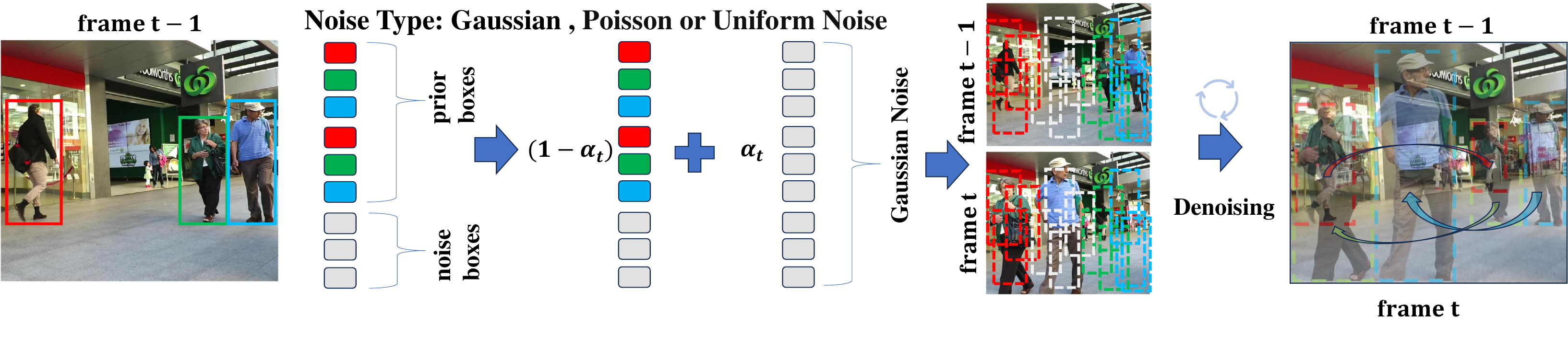}
  \caption{The inference of DiffusionTrack can be divided into three steps: (1) padding repeated prior boxes with given noise boxes until predefined number $N_{test}$ is reached. (2) adding Gaussian noise to input boxes according to $\mathbf{B}=(1-\alpha_{t}) \cdot \mathbf{B}+\alpha_{t} \cdot \mathbf{B}_{noise}$ under the control of $\alpha_{t}$. (3) getting tracking results by a denoising process with the number of DDIM sampling steps $s$.
}
  \label{fig3}
\end{figure*}

\section{Method}
In this section, we present our DiffusionTrack. In contrast to existing motion-based and query-based methods, we design a consistent tracker that performs tracking implicitly by predicting and associating the same object across two adjacent frames within the video sequence. We first briefly review the pipeline of multi-object tracking and diffusion models. Then, we introduce the architecture of DiffusionTrack. Finally, we present model training and inference.

\subsection{Preliminaries}

\textbf{Multi-object tracking}. The learning objective of MOT is a set of input-target pairs $(\mathbf{X}_{t}, \mathbf{B}_{t}, \mathbf{C}_{t})$ sorted by time $t$, where $\mathbf{X}_{t}$ is the input image at time $t$, $\mathbf{B}_{t}$ and $\mathbf{C}_{t}$ are a set of bounding boxes and category labels for objects in the video at time $t$ respectively. More specifically, we formulate the $i$-th box in the set $\mathbf{B}_{t}$ as $\mathbf{B}_{t}^{i}$ = $(c^{i}_{x}, c^{i}_{y}, w_{i}, h_{i})$, where $(c^{i}_{x}, c^{i}_{y})$ is the center coordinates of the bounding box, $(w_{i}, h_{i})$ are width and height of that bounding box, $i$ is the identity number respectively. Specially, $\mathbf{B}_{t}^{i}$ = $\mathbf{\emptyset}$ when $i$-th object miss in $\mathbf{X}_{t}$.

\noindent\textbf{Diffusion model}. Recent diffusion models usually use two Markov chains: a forward chain that perturbs the image to noise and a reverse chain that refines noise back to the image. Formally, given a data distribution $\mathbf{x}_0 \sim q(\mathbf{x}_0)$, the forward noise perturbing process at time $t$ is defined as $q(\mathbf{x}_t|\mathbf{x}_{t-1})$. It gradually adds Gaussian noise to the data according to a variance schedule $\beta_1,\cdots,\beta_T$:
\begin{equation}
q(\mathbf{x}_t|\mathbf{x}_{t-1}) = \mathcal{N}(\mathbf{x}_t;\sqrt{1-\beta_t}\mathbf{x}_{t-1},\beta_t\mathbf{I}).
\end{equation}
Given $\mathbf{x}_0$, we can easily obtain a sample of $\mathbf{x}_t$ by sampling a Gaussian vector $\mathbf{\epsilon} \sim \mathcal{N}(\mathbf{0},\mathbf{I})$ and applying the transformation as follows:
\begin{equation}
\mathbf{x}_t = \sqrt{\bar{\alpha_t}}\mathbf{x}_0+(1-\bar{\alpha_t})\mathbf{\epsilon},
\end{equation}
where $\bar{\alpha_t} = \prod^t_{s=0}(1-\beta_s)$.
During training, a neural network predict $\mathbf{x}_0$ from $\mathbf{x}_t$ for different $t\in \{1,\cdots,T\}$. In inference, we start from a random noise $\mathbf{x}_T$ and iteratively apply the reverse chain to obtain $\mathbf{x}_0$.

\subsection{DiffusionTrack}
The overall framework of our DiffusionTrack is visualized in Figure \ref{fig2}, which consists of two major components: a feature extraction backbone and a data association denoising head (diffusion head), where the former runs only once to extract a deep feature representation from two adjacent input image $(\mathbf{X}_{t-1}, \mathbf{X}_{t})$, and the latter takes this deep features as condition, instead of two adjacent raw images, to progressively refine the paired association box predictions from paired noise boxes.
In our setting, data samples are a set of paired bounding boxes $\mathbf{z}_0 = (\mathbf{B}_{t-1},\mathbf{B}_{t})$, where $\mathbf{z}_0\in \mathbf{R}^{N \times 8}$. A neural network $f_\theta(\mathbf{z}_s, s, \mathbf{X}_{t-1},\mathbf{X}_{t}) \quad s=\{0,\cdots,T \}$ is trained to predict $\mathbf{z}_0$ from paired noise boxes $\mathbf{z}_s$, conditioned on the corresponding two adjacent images $(\mathbf{X_{t-1}},\mathbf{X_{t}})$. The corresponding category label $(\mathbf{C}_{t-1},\mathbf{C}_{t})$ and association confidence score $\mathbf{S}$ are produced accordingly. If $\mathbf{X_{t-1}}=\mathbf{X_{t}}$, the multi-object tracking task degenerates into an object detection problem. The consistent design allows DiffusionTrack to solve the two tasks simultaneously.

\noindent\textbf{Backbone}. We employ the backbone of YOLOX ~\cite{YOLOX}
as our backbone. The backbone extracts high-level features of the two adjacent frames with FPN ~\cite{FPN} and then feeds them into the following diffusion head for conditioned data association denoising.

\noindent\textbf{Diffusion head}. The diffusion head takes a set of proposal boxes as input to crop RoI-feature ~\cite{prroipooling} from the feature map generated by the backbone and sends these RoI-features to different blocks to obtain box regression, classification results, and association confidence scores, respectively. To solve the object tracking problem, we add a spatial-temporal fusion module (STF) and an association score head to each block of the diffusion head.

\noindent\textbf{Spatial-temporal fusion module}. We design a new spatial-temporal fusion module so that the same paired box can exchange temporal information with each other to ensure that the data association on two consecutive frames can be completed. Given the RoI-features $\mathbf{f}^{t-1}_{roi}$, $\mathbf{f}^{t}_{roi}\in\mathbb{R}^{N\times R \times d}$, and the self-attention output query $\mathbf{q}^{t-1}_{pro}$, $\mathbf{q}^{t}_{pro}\in\mathbb{R}^{N\times d}$ at current block, we conduct linear project and batch matrix multiplication to get the object query $\mathbf{q}^{t-1}$, $\mathbf{q}^{t}\in\mathbb{R}^{N\times d}$ as:
\begin{equation}
\begin{split}
  & \mathbf{P}^{i}_{1}, \mathbf{P}^{i}_{2} = \mathbf{Split}(\mathbf{Linear1}(\mathbf{q}^{i}_{pro})), \\
    &\mathbf{feat}=\mathbf{Bmm}(\mathbf{Bmm}(\mathbf{Concat}(\mathbf{f}^{i}_{roi},\mathbf{f}^{j}_{roi}),\mathbf{P}^{i}_{1}),\mathbf{P}^{i}_{2})\\
    &\mathbf{q}^{i}=\mathbf{Linear2}(\mathbf{feat}),
  \quad \mathbf{q}^{i}\in\mathbb{R}^{N\times d} \\
  &(i,j) \in\left[(t-1, t),(t,t-1)\right]
  \label{eq:1}
\end{split}
\end{equation}


\noindent\textbf{Association score head}. In addition to the box head and class head, we add an extra association score head to obtain the confidence score of the data association by feeding the fused features of the two paired boxes into a Linear Layer. The head is used to determine whether the paired boxes output belongs to the same object in the subsequent Non-Maximum Suppression (NMS) post-processing process.

\subsection{Model Training and Inference}
In the training phase, our approach takes a pair of frames randomly sampled from sequences in the training set with an interval of 5 as input. we first pad some extra boxes to original ground-truth boxes appearing in both frames such that all boxes are summed up to a fixed number $N_{train}$. Then we add Gaussian noise to the padded ground-truth boxes with the monotonically decreasing cosine schedule for $\alpha_{t}$ in time step $t$. We finally conduct a denoising process to get association results from these constructed noise boxes. We also design a baseline that only corrupts the ground-truth boxes in frame $t$ and conditionally denoises the corrupted boxes based on the prior boxes in frame $t-1$ to verify the necessity of corruption design for both frames in DiffusionTrack.

\noindent \textbf{Loss Function}. GIoU ~\cite{giou} loss is an extension of IoU loss which solves the problem that there is no supervisory information when the predicted boxes have no intersection with the ground-truth. We extend the definition of GIoU to make it compatible with paired boxes design. 3D GIoU and 3D IoU are the volume-extended versions of the original area ones. For each pair paired $(\mathbf{T}_{d},\mathbf{T}_{gt})$ in the matching set M obtained by the Hungarian matching algorithm, we denote its class score, predicted boxes result, and association score as $(\mathbf{C}_{d}^{t-1},\mathbf{C}_{d}^{t})$,
$(\mathbf{B}_{d}^{t-1},\mathbf{B}_{d}^{t})$, and $\mathbf{S}_{d}$. The training loss function can be formulated as:
\begin{equation}
\begin{split}
 & \mathcal{L}_{cls}(\mathbf{T}_{d},\mathbf{T}_{gt})= \sum_{i=t-1}^{t} {\mathcal{L}_{cls}(\sqrt{\mathbf{C}_{d}^{i}\times \mathbf{S}_{d}},\mathbf{C}_{gt}^{i})} \\
  & \mathcal{L}_{reg}(\mathbf{T}_{d},\mathbf{T}_{gt})= \sum_{i=t-1}^{t} {\mathcal{L}_{reg}(\mathbf{B}_{d}^{i},\mathbf{B}_{gt}^{i})}
 \\
 & \mathcal{L}_{det} =  \frac{1}{N_{pos}} \sum_{(\mathbf{T}_{d},\mathbf{T}_{gt}) \in \mathbf{M}}{\lambda_{1}\mathcal{L}_{cls}(\mathbf{T}_{d},\mathbf{T}_{gt}) \quad + }
 \\ & {\lambda_{2} \mathcal{L}_{reg}(\mathbf{T}_{d},\mathbf{T}_{gt}) + \lambda_{3}(1-GIoU_{3d}(\mathbf{T}_{d},\mathbf{T}_{gt}))}
\end{split}
  \label{eq:2}
\end{equation}
\noindent where $\mathbf{T}_{d}$ and $\mathbf{T}_{gt}$ are square frustums consisting of estimated detection boxes and ground-truth bounding boxes for the same target in two adjacent frames respectively. $N_{pos}$ denotes the number of positive foreground samples. $\lambda_{1}$, $\lambda_{2}$ and $\lambda_{3}$
are the weight coefficients that are assigned as 2, 5 and 2 during training experiments. $\mathcal{L}_{cls}$ is the focal loss proposed in ~\cite{FocalLoss} and $\mathcal{L}_{reg}$
is the $L_{1}$ loss.

As shown in Figure.\ref{fig3}, the inference pipeline of DiffusionTrack is a denoising sampling process from paired noise boxes to association results. Unlike the detection task that selects random boxes from the Gaussian distribution, the tracking task has prior information about an object in the frame $t-1$, so we can use prior boxes to generate initialized noise boxes with a fixed number of $N_{test}$ as in the training phase to benefit data association. In contrast to DiffusionTrack, we simply repeat the prior box without padding extra random boxes and add Gaussian noise to prior boxes only at $t$ in the baseline model. Once the association results are derived, IoU is utilized as the similarity metric to connect the object tracklets. To address potential occlusions, a simple Kalman filter is implemented to reassociate lost objects and more details exist in the Appendix. 

\begin{figure}[t]
\centering
\subfloat[Dynamic boxes and progressive refinement. DiffusionTrack is trained on the MOT17 train-half set with 500 proposal boxes and evaluated on the MOT17 val-half set with different numbers of proposal boxes. More sampling steps and proposal boxes in inference bring performance gain, but the effect is gradually saturated]
{
\hspace{-8mm}
\includegraphics[width=0.98\linewidth]{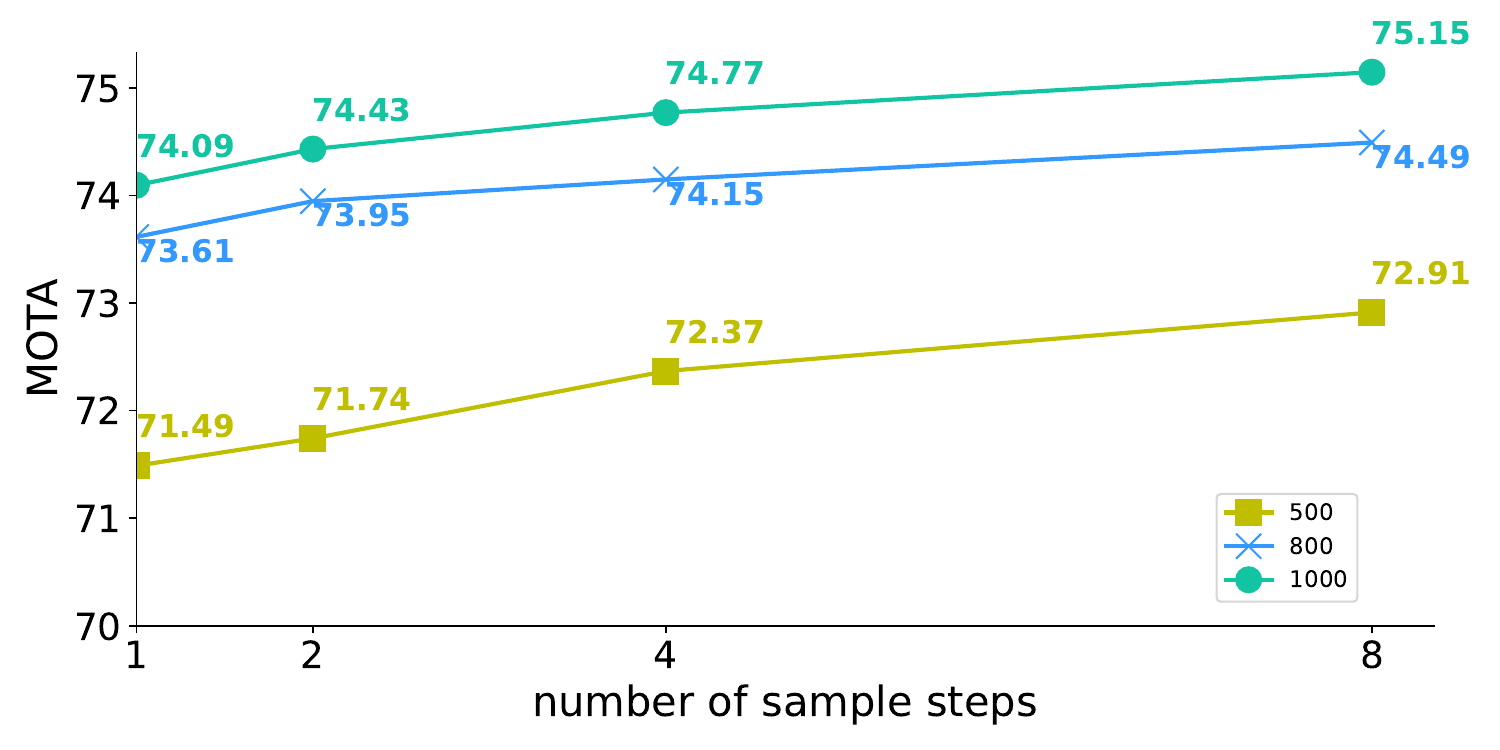}
\label{fig:progressive}
}
\\
\subfloat[Robustness to detection perturbation. All trackers are trained on MOT17 training set and evaluated on MOT17 val-half set with little detection perturbation as $\mathbf{B}_{det}=(1-\alpha_{t})\cdot \mathbf{B}_{det}+\alpha_{t}\cdot \mathbf{B}_{noise}$. DiffusionTrack is robust to perturbation attacks with 800 proposal boxes while other approaches are vulnerable.]
{
\hspace{-8mm}
\includegraphics[width=0.98\linewidth]{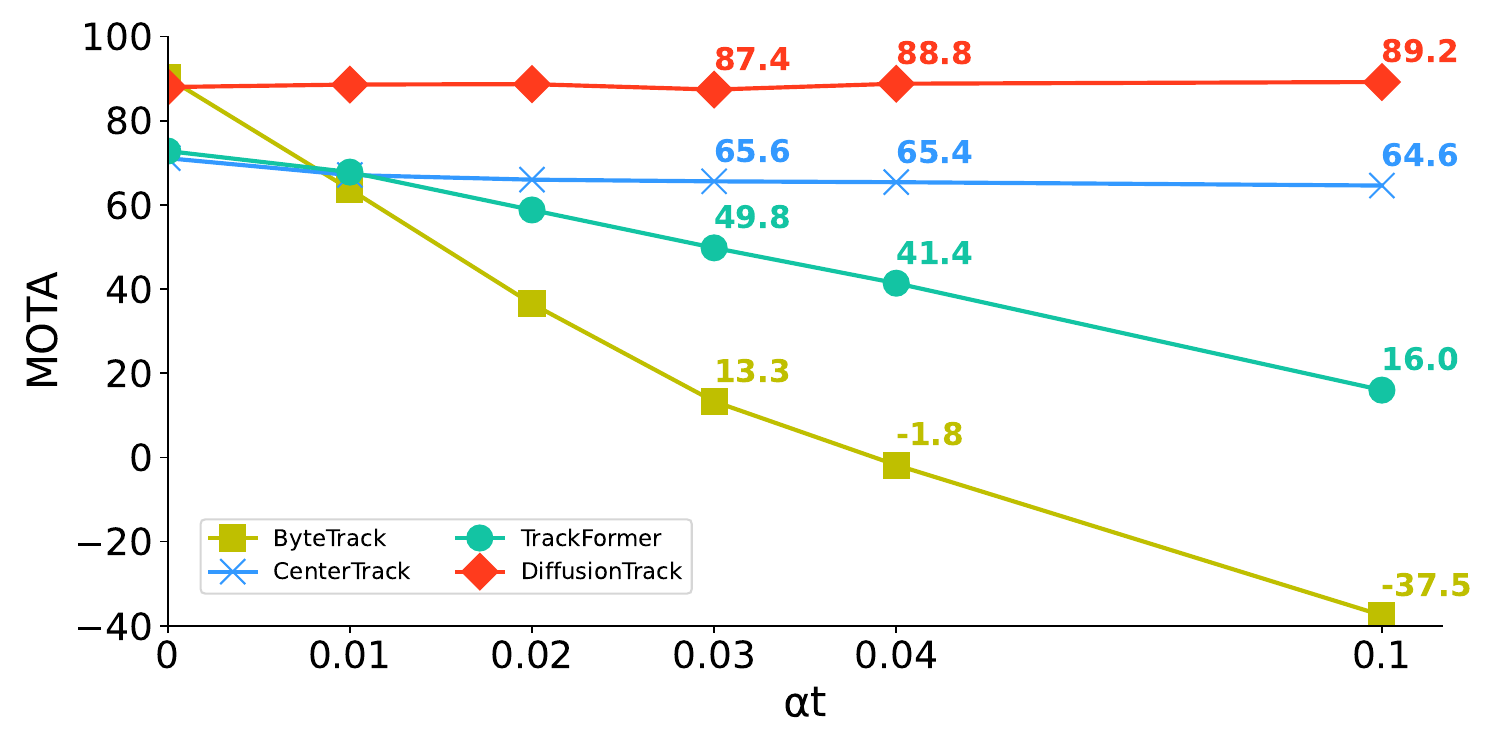}
\label{fig:robustness}
}
\caption{Intriguing properties of DiffusionTrack. DiffusionTrack obtains performance gain by enlarging proposal box numbers and sampling steps while being robust to detection perturbation compared with the previous tracker.}
\label{fig:intriguing properties}
 
\end{figure}
\section{Experiments}
In this section, we first introduce experimental setting and show the intriguing properties of DiffusionTrack. Then we verify the individual contributions in the ablation study and finally present the tracking evaluation on several challenging benchmarks, including MOT17~\cite{MOT16}, MOT20~\cite{MOT20} and DanceTrack~\cite{DanceTrack}. 
We also present the comparison with baseline model and carry out a deep analysis for DiffusionTrack.

\begin{table*}[!htpb] \small
\centering
\subfloat[
Proportion of prior information. Using prior information benefit data association.\label{tab:prior_info}
]{
\centering
\begin{minipage}{0.44\linewidth}{
\begin{center}
\begin{tabular}{l|cccc}
\toprule
prior Info & \multirow{2}{*}{MOTA} & \multirow{2}{*}{IDF1} & \multirow{2}{*}{HOTA} & \multirow{2}{*}{AssA} \\
proportion & & & &  \\
\midrule[0.5pt]
0\% & 71.2 & 65.9 & 58.1 & 54.9\\
\rowcolor{gray!40}
25\% & 73.6 & 70.0 & 60.7 & 58.4\\
50\% & \textbf{74.5} & 71.2 & 61.8 & 60.1\\
75\% & 74.1 & \textbf{71.4} & \textbf{61.9} & \textbf{60.7}\\
100\% & 72.9 & 66.8 & 58.4 & 54.7\\
\bottomrule
\end{tabular}
\end{center}}
\end{minipage}
}
\hspace{1em}
\subfloat[Box padding strategy. Compared to other padding strategy, concatenating Gaussian noise works best.\label{tab:padding stragety}
]{
\centering
\begin{minipage}{0.5 \linewidth}{\begin{center}
\begin{tabular}{l|cccc}
\toprule
padding & \multirow{2}{*}{MOTA} & \multirow{2}{*}{IDF1} & \multirow{2}{*}{HOTA} & \multirow{2}{*}{AssA} \\
strategy & &  &  &  \\
\midrule[0.5pt]
Repeat & 72.9 & 66.8 & 58.4 & 54.7 \\
Cat Poisson & 71.9 & 67.1 & 58.9 & 56.1\\
\rowcolor{gray!40}
Cat Gaussian & \textbf{73.6} & \textbf{70.0} &\textbf{60.7} & \textbf{58.4}\\
Cat Uniform & 71.5 & 63.9 & 56.8 & 52.2\\
Cat Full & 71.2 & 64.4 & 57.3 & 53.7\\
\bottomrule
\end{tabular}
\end{center}}
\end{minipage}
}
\\
\subfloat[
Perturbation schedule. Choosing $t$ through a logarithmic perturbation strategy works best.\label{tab:noise_schedule}]{
\begin{minipage}{0.44 \linewidth}{
\begin{center}
\begin{tabular}{l|cccc}
\toprule
perturbation & \multirow{2}{*}{MOTA} & \multirow{2}{*}{IDF1} & \multirow{2}{*}{HOTA} & \multirow{2}{*}{AssA} \\
strategy $f(x)$ & & & &  \\
\midrule[0.5pt]
$0.4$ & 73.0 & 67.2 & 58.2 & 54.2 \\
\rowcolor{gray!40}
$x$ &  73.6 & 70.0 & 60.7 & 58.4\\
$(e^x-1)/(e-1)$ & 74.3 & 70.5 & 61.4 & 59.7 \\
$log(x+1)/log2$ & \textbf{74.4} & \textbf{72.0} & \textbf{62.6} & \textbf{61.9}\\
\bottomrule
\end{tabular}
\end{center}
}
\end{minipage}
}
\hspace{1em}
\subfloat[Efficiency comparison. Adopting more proposal boxes and sampling steps brings performance gain at the cost of latency. \label{tab:efficiency}
]{
\begin{minipage}{0.50\linewidth}{\begin{center}
\begin{tabular}{lc|cccccc}
\toprule
\multirow{2}{*}{box} & sampling & \multirow{2}{*}{MOTA} & \multirow{2}{*}{IDF1} & \multirow{2}{*}{HOTA} & \multirow{2}{*}{FLOPs(G)} & \multirow{2}{*}{FPS} \\
& step & & & & \\
\midrule[0.5pt]
500  & 1 & 71.5 & 66.3 & 58.4 & \textbf{229.6} &\textbf{21.05} \\
500  & 2 & 71.7 & 68.1 & 59.5 & 459.2 & 10.47 \\
\rowcolor{gray!40}
800  & 1 & 73.6 & 70.0 & 60.7 & 367.3 & 15.89 \\
1000  & 1 & \textbf{74.1} & \textbf{70.7} & \textbf{61.3} & 459.1 & 13.37\\
\bottomrule
\end{tabular}
\end{center}}
\end{minipage}
}
\caption{Ablation experiments. The model is trained on the MOT17 train-half and tested on the MOT17 val-half. Default settings are marked in gray. See Sec \ref{sec:ablation} for more details.}
\end{table*}

\subsection{Setting}

\noindent \textbf{Datasets.} We evaluate our method on multiple multi-object tracking datasets including MOT17 \cite{MOT16}, MOT20 \cite{MOT20} and DanceTrack~\cite{DanceTrack}. MOT17 and MOT20 are for pedestrian tracking, where targets mostly move linearly, while scenes in MOT20 are more crowded. For the data in DanceTrack, the objects have a similar appearance, severe occlusion, and frequent crossovers with highly non-linear motion. 

\noindent \textbf{Metric.} We mainly use Multiple Object Tracking Accuracy (MOTA) ~\cite{mota}, Identity F1 Score (IDF1) ~\cite{idf1}, and Higher Order Tracking Accuracy (HOTA) ~\cite{hota} for evaluation. 

\noindent \textbf{Implementation Details}. We adopt the pre-trained YOLOX detector from ByteTrack \cite{ByteTrack} and train DiffusionTrack on MOT17, MOT20, and DanceTrack training sets in two phases. For MOT17, the training schedule consists of 30 epochs on the combination of MOT17, CrowdHuman, Cityperson and ETHZ for detection and another 30 epochs on MOT17 solely for tracking. For MOT20, we only add CrowdHuman as additional training data. For DanceTrack, we do not use additional training data and only train 40 epochs. We also use Mosaic \cite{bochkovskiy2020yolov4} and Mixup \cite{zhang2017mixup} data augmentation during the detection and tracking training phases. The training samples are directly sampled from the same video within the interval length of 5 frames. The size of an input image is resized to 1440$\times$800. The 236M trainable diffusion head parameters are initialized with Xavier Uniform. The AdamW~\cite{adamw} optimizer is employed with an initial learning rate of 1e-4, and the learning rate decreases according to the cosine function with the final decrease factor of 0.1. We adopt a warm-up learning rate of 2.5e-5 with a 0.2 warm-up factor on the first 5 epochs. We train our model on 8 NVIDIA GeForce RTX 3090 with FP32-precision and a constant seed for all experiments. The mini-batch size is set to 16, with each GPU hosting two batches with $N_{train}=500$. Our approach is implemented in Python 3.8 with PyTorch 1.10. We set association score threshold $\mathbf{\tau}_{conf}=0.25$, 3D NMS threshold $\mathbf{\tau}_{nms3d}=0.6$, detection score threshold $\mathbf{\tau}_{det}=0.7$ and 2D NMS threshold $\mathbf{\tau}_{nms2d}=0.7$ for default hyper-parameter setting. The total training time
is about 30 hours, and FPS is measured with FP16-precision and batch size of 1 on a single GPU.

\subsection{Intriguing Properties}
DiffusionTrack has several intriguing properties, such as the ability to achieve better accuracy through more boxes or/and more refining steps at the higher latency cost, and strong robustness to detection perturbation for safety application.

\noindent \textbf{Dynamic boxes and progressive refinement}. Once the model is trained, it can be used by changing the number of boxes and the number of sample steps in inference. Therefore, we can deploy a single DiffusionTrack to multiple scenes and obtain a desired speed-accuracy trade-off without retraining the network. In Figure \ref{fig:progressive}, we evaluate DiffusionTrack with 500, 800, and 1000 proposal boxes by increasing their sampling steps from 1 to 8, showing that high MOTA in DiffusionTrack could be achieved by either increasing the number of random boxes or the sampling steps.

\noindent \textbf{Robustness to detection perturbation}. Almost all previous approaches are very sensitive to detection perturbation which poses significant risks to safety-critical applications such as autonomous driving. Figure \ref{fig:robustness} shows the robustness of the four mainstream trackers under detection perturbation. As can be seen from the performance comparison, DiffusionTrack has no performance penalty for perturbation, while other trackers are severely affected, especially the two-stage ByteTrack.

\begin{table*}[!htbp]\normalsize
  \centering
  \setlength{\tabcolsep}{1.5mm}{
    \begin{tabular}{@{ }l|c@{ }c@{ }c@{ }c@{ }c@{ }c@{ }c || c@{ }c@{ }c@{ }c@{ }c@{ }c@{ }c@{ }}
    \toprule
    & \multicolumn{7}{c||}{MOT17} & \multicolumn{7}{c}{MOT20} \\
    \midrule[0.5pt]
    Methods & MOTA$\uparrow$ & IDF1$\uparrow$ & HOTA$\uparrow$ & AssA$\uparrow$ & DetA$\uparrow$ & IDs$\downarrow$ & Frag$\downarrow$ & MOTA$\uparrow$ & IDF1$\uparrow$ & HOTA$\uparrow$ & AssA$\uparrow$ & DetA$\uparrow$ & IDs$\downarrow$ & Frag$\downarrow$\\
    \midrule[0.5pt]
    \textit{Two-Stage:}\\
    OC-SORT& 78.0 & 77.5 & 63.2 & 63.4 & 63.2 & 1950 & 2040 & 75.7 & 76.3 & 62.4 & 62.5 & 62.4 & 942 & 1086\\
    BoT-SORT& 80.5 & \textbf{80.2} & \textbf{65.0} & \textbf{65.5} & \textbf{64.9} & 1212 & \textbf{1803} & 77.8 & \textbf{77.5} & \textbf{63.3} & 62.9 & \textbf{64.0} & 1313 & 1545\\
    Bytetrack& 80.3 & 77.3 & 63.1 & 62.0 & 64.5 & 2196 & 2277 & 77.8 & 75.2 & 61.3 & 59.6 & 63.4 & 1223 & 1460\\
    StrongSORT& 79.6 & 79.5 & 64.4 & 64.4 & 64.6 & \textbf{1194} & 1866 & 73.8 & 77.0 & 62.6 & \textbf{64.0} & 61.3 & \textbf{770} & \textbf{1003}\\
    P3AFormer& \textbf{81.2} & 78.1 & / & / & / & 1893 & / & \textbf{78.1} & 76.4 & / & / & / & 1332 & / \\
    GMTracker& 61.5 & 66.9 & / & / & / & 2415 & / & / & / & / & / & / & / & / \\
    GNMOT& 50.2 & 47.0 & / & / & / & 5273 & / & / & 76.4 & / & / & / & / & / \\
    \midrule[0.5pt]
    \textit{One-Stage:}\\
    TrackFormer& 74.1 & 68.0 & 57.3 & 54.1 & 60.9 & 2829 & 4221 & 68.6 & 65.7 & 54.7 & 53.0 & 56.7 & \textbf{1532} & \textbf{2474}\\
    MeMOT& 72.5 & 69.0 & 56.9 & 55.2 & / & 2724 & / & 63.7 & 66.1 & 54.1 & \textbf{55.0} & / & 1938 & / \\
    MOTR& 71.9 & 68.4 & 57.2 & 55.8 & / & \textbf{2115} & \textbf{3897} &/&/&/&/&/&/&/ \\
    CenterTrack& 67.8 & 64.7 & 52.2 & 51.0 & 53.8 & 3039 & 6102 &/&/&/&/&/&/&/ \\
    PermaTrack& 73.8 & 68.9 & 55.5 & 53.1 & 58.5 & 3699 & 6132 &/&/&/&/&/&/&/ \\
    TransCenter& 73.2 & 62.2& 54.5 & 49.7 & 60.1 & 4614 & 9519 & 67.7 & 58.7 & / & / & / & 3759 & / \\
    \underline{GTR}& \underline{75.3} & \underline{71.5} & \underline{59.1} & \underline{57.0} & \underline{61.6} & \underline{2859} & / & / & / & / & / & / & / & / \\
    TubeTK& 63.0 & 58.6 & / & / & / & 4137 & / & / & / & / & / & / & / & / \\
    \textbf{Baseline}& 74.6 & 66.7 & 55.9 & 50.8 & 61.9 & 16375 & 7206 & 63.3 & 49.5 & 42.5 & 34.7 & 52.5 & 9990 & 6710 \\
    \textbf{DiffusionTrack}& \textbf{77.9} & \textbf{73.8} & \textbf{60.8} & \textbf{58.8} & \textbf{63.2} & 3819 & 4815 & \textbf{72.8} & \textbf{66.3} & \textbf{55.3} & 51.3 & \textbf{59.9} & 4117 & 4446\\
    \bottomrule
    \end{tabular}
  }
\caption{Performance comparison to state-of-the-art approaches on the MOT17 and MOT20 test set with the private detections. The best results are shown in bold. The offline method is marked in underline.}
  \label{tab:sota}
  \vspace{-3pt}
\end{table*}


\begin{table}[!htpb]\small
  \centering
  \setlength{\tabcolsep}{1mm}{
    \begin{tabular}{l@{ }|ccccc}
    \toprule
    Methods & HOTA$\uparrow$ & MOTA $\uparrow$ & DetA$ \uparrow$ & AssA $\uparrow$ & IDF1$\uparrow$ \\
    \midrule[0.5pt]
    QDTrack& 45.7 & 83.0 & 72.1 & 29.2 & 44.8\\
    TraDes& 43.3 & 86.2 & 74.5 & 25.4 & 41.2\\
    SORT& 47.9 & \textbf{91.8} & 72.0 & 31.2 & 50.8\\
    ByteTrack& 47.3 & 89.5 & 71.6 & 31.4 & 52.5\\
    OC-SORT & \textbf{54.6} & 89.6 & \textbf{80.4} & \textbf{40.2} & \textbf{54.6}\\
    \midrule[0.5pt]
    TransTrack & 45.5 & 88.4 & 75.9 & 27.5 & 45.2\\
    CenterTrack & 41.8 & 86.8 & 78.1 & 22.6 & 35.7\\
    \underline{GTR} & \underline{48.0} & \underline{84.7} & \underline{72.5} & \underline{31.9} & \underline{50.3}\\
    \textbf{Baseline} & 44.0 & 79.4 & 74.1 & 26.2 & 40.2\\
    \textbf{DiffusionTrack} & \textbf{52.4} & \textbf{89.3} & \textbf{82.2} & \textbf{33.5} & \textbf{47.5}\\
    \bottomrule
    \end{tabular}
  }
\caption{Performance comparison to state-of-the-art approaches on the DanceTrack test set. The best results are shown in bold. Offline method is marked in underline}
  \label{tab:DanceTrack}
\end{table}



\subsection{Ablation Study}
\label{sec:ablation} 
We conduct ablation experiments on several relevant factors in Figure \ref{fig3} to study DiffusionTrack in detail.

\noindent\textbf{Proportion of prior information}.
In contrast to object detection, multi-object tracking has prior information about the object location in the previous frame $t-1$. When constructing $N_{test}$ proposal boxes, we can control the proportion of prior information by simply repeating prior boxes. we can find that an appropriate proportion of prior information can improve the tracking performance from Table \ref{tab:prior_info}.

\noindent\textbf{Box padding strategy}.
Table \ref{tab:padding stragety} shows different box padding strategies. Our Concatenating Gaussian random boxes outperforms repeating existing prior boxes, concatenating random boxes in different noise types or image-size.

\noindent\textbf{Perturbation schedule}. Proposal boxes are initialized by adding Gaussian noise to padded prior boxes under the control of $\alpha_{t}$. We need a perturbation schedule to deal with complicated scenes, such as a larger $\alpha_{t}$ when facing non-linear object motion. The perturbation schedule can be modeled by $t$ and formulated as $t=1000\cdot f(x)$, where $x$ is the average percentage of object motion cross two frames and $f$ is the perturbation schedule function. As shown in Table \ref{tab:noise_schedule}, using a logarithmic function $f(x)=\frac{log(x+1)}{log2}$ as perturbation schedule works best.

\noindent\textbf{Efficiency comparison}. Table \ref{tab:efficiency} shows the efficiency comparison with different numbers of proposal boxes and sampling steps. The run time is evaluated on a single NVIDIA GeForce 3090 GPU with a mini-batch size of 1 and FP16-precision. We observe that more refinements cost brings more performance gain and results in less FPS. DiffusionTrack can flexibly choose different settings for every single frame to deal with complicated scenes within a video.

\subsection{State-of-the-art Comparison}

Here we report the benchmark results of DiffusionTrack and baseline compared with other mainstream methods on multiple datasets. We evaluated DiffusionTrack on DanceTrack, MOT17, and MOT20 test datasets with 500, 800, and 1000 noise boxes respectively in same default setting. 

\noindent \textbf{MOT17 and MOT20}. We use the standard split and obtain the test set evaluation by submitting the results to the online website. As can be seen from the performance comparison in Table\ref{tab:sota}, our DiffusionTrack achieves state-of-the-art both in MOT17 and MOT20 for one-stage methods with the MOTA of 77.9 and 72.8 respectively. 

\noindent \textbf{DanceTrack}. To evaluate DiffusionTrack under challenging non-linear object motion, we report results on the DanceTrack in Table ~\ref{tab:DanceTrack}. DiffusionTrack achieves the state-of-the-art on DanceTrack with HOTA (52.4). 

The baseline model has a close performance to DiffusionTrack on MOT17 but performs very poorly on MOT20 and DanceTrack. In our understanding, Baseline simply learns a coordinate regression between boxes $\mathbf{B}_{t-1}$ and boxes $\mathbf{B}_{t}$ at conditioned on the pooled features at time $t-1$ which can not deal with crowed and non-linear object motion problem. We guess the coarse-to-fine diffusion process is a special data-augmented method that can enable DiffusionTrack to discriminate between various objects.

\section{Conclusion}
In this work, we propose a novel end-to-end multi-object tracking approach that formulates object detection and association jointly as a consistent denoising diffusion process from paired noise boxes to object association. Our noise-to-tracking pipeline has several appealing properties, such as dynamic box and progressive refinement, consistent model structure, and robustness to perturbation detection results, enabling us to to obtain the desired speed-accuracy trade-off with same network parameters. Extensive experiments show that DiffusionTrack achieves favorable performance compared to previous strong baseline methods. We hope that our work will provide a interesting insight into multi-object tracking from the perspective of the diffusion model, and that the performance of a wide variety of trackers can be enhanced by local or global denoising processes.

\clearpage

\appendix

\section*{\centering Appendix}
\
\newline In this supplementary material, we describe limitation and broader impact of our method in Section \ref{Limitations and Broader Impacts}. And we presents the details about 3D GIoU calculation in Section \ref{3D GIoU}. In Section \ref{Visualization}, we illustrate extensive visual results of the non-linear motion and crowded scenes in Dancetrack~\cite{DanceTrack} and MOT20~\cite{MOT20} sequences. Finally, we show the Pseudo-code of DiffusionTrack in Section \ref{Inference Detail}

\section{Limitation and Broader Impact}
\label{Limitations and Broader Impacts}
\noindent\textbf{Limitation.} Although our proposed DiffuionTrack can perform detection and tracking jointly through a progressive denoising process. We observe that our tracker does not work well for small objects on MOT20 due to the weakness of the diffusion model. The unsatisfactory performance gained from multi-step denoising and the longer training time is also intolerable. In the future, we plan to adopt more advanced diffusion models and efficient attention module to reduce the time spent on model training and inference.

\noindent\textbf{Broader Impact.} From the results presented by DiffusionTrack, we believe the diffusion model is a new possible solution to MOT due to its simple training and inference pipeline with consistent model design, showing great potential for high-level semantic correlation of objects. We hope our work could serve as a simple yet effective baseline, which could inspire designing more efficient frameworks and rethinking the learning objective for the challenging MOT task.

\section{3D GIoU Calculation}
\label{3D GIoU}
\begin{figure}[htbp]
  \centering
  \includegraphics[width=0.96\linewidth,keepaspectratio]{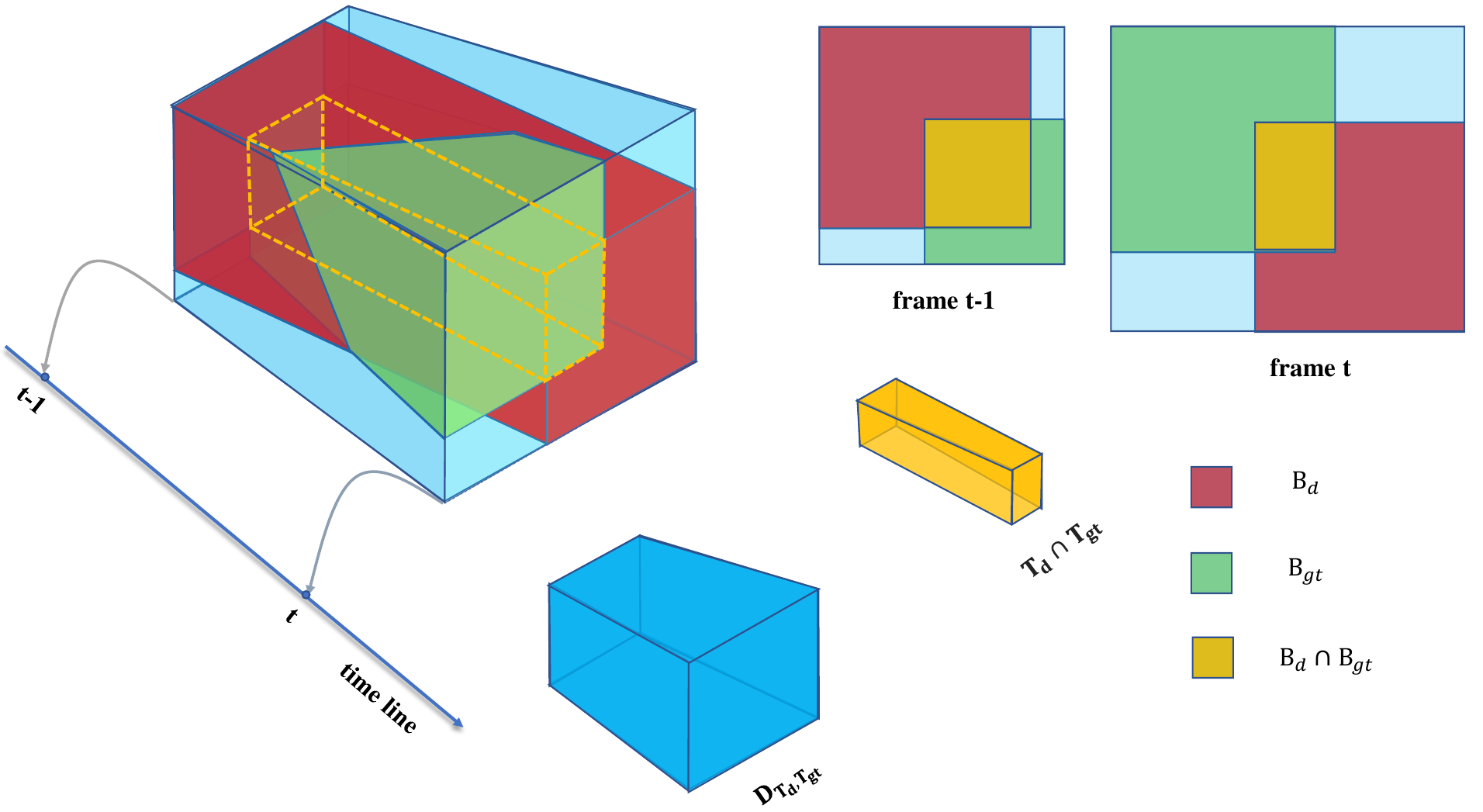}
  \caption{Visualization of the calculation process of 3D GIoU. 3D GIoU and 3D IoU are the volume extended version of the original area ones. The intersection $T_{d}\cap T_{gt}$ and $D_{T_{d},T_{gt}}$ of targets between two adjacent frames are square frustums, thus the volume of them can be calculated in the same way of original GIoU.}
  \label{fig:giou3d}
\end{figure}

 IoU is the most popular indicator to evaluate the quality of the predicted Bbox, and it is usually used as the loss function. GIoU ~\cite{giou} loss is an extension of IoU loss which solves the problem that there is no supervisory information when the predicted boxes have no intersection with the ground truth. We extend the definition of GIoU to make it compatible with paired boxes design. 3D GIoU of paired predicted boxes is defined as. As shown in Figure \ref{fig:giou3d}, 3D GIoU of paired predicted boxes is defined as:

\begin{equation*}
\begin{split}
  & IoU_{3D}(\mathbf{T}_{d},\mathbf{T}_{gt}) 
  = 
 \frac{\sum_{i=t-1}^{t}Area(\mathbf{B}_{d}^{i}\cap \mathbf{B}_{gt}^{i})}{\sum_{i=t-1}^{t}Area(\mathbf{B}_{d}^{i}\cup \mathbf{B}_{gt}^{i})} \\
    & GIoU_{3D}(\mathbf{T}_{d},\mathbf{T}_{gt}) = IoU_{3D}(\mathbf{T}_{d},\mathbf{T}_{gt}) \quad - \\ 
    & \frac{|\sum_{i=t-1}^{t} Area(\mathbf{D}_{\mathbf{B}_{d}^{i},\mathbf{B}_{gt}^{i}}) - Area(\mathbf{B}_{d}^{i}\cup \mathbf{B}_{gt}^{i})|}{|\sum_{i=t-1}^{t} Area(\mathbf{D}_{\mathbf{B}_{d}^{i},\mathbf{B}_{gt}^{i}})|}
\end{split}
\label{eq:3}
\end{equation*}
\noindent Where $\mathbf{D}_{\mathbf{B}_{d}^{i},\mathbf{B}_{gt}^{i}}$ is the smallest enclosing convex object of estimated detection box $\mathbf{B}_{d}$ and ground-truth bounding box $\mathbf{B}_{gt}$ at frame $i$. $\mathbf{T}_{d}$ and $\mathbf{T}_{gt}$ are square frustums consisting of estimated detection boxes and ground-truth bounding boxes for the same target in two adjacent frames $t-1$,$t$, respectively. Similarly, the intersection $\mathbf{T}_{d}\cap \mathbf{T}_{gt}$ is also a square frustum that consists of the intersection $\mathbf{B}_{d}^{t-1}\cap B_{gt}^{t-1}$ and the intersection $\mathbf{B}_{d}^{t}\cap \mathbf{B}_{gt}^{t}$. 3D GIoU and 3D IoU are the volume-extended versions of the original area ones.

\begin{figure}[h]
	\centering
        \subfloat[Crowded Scene]{
	\begin{minipage}[t]{1\linewidth}
            \includegraphics[width=0.98\columnwidth,keepaspectratio]{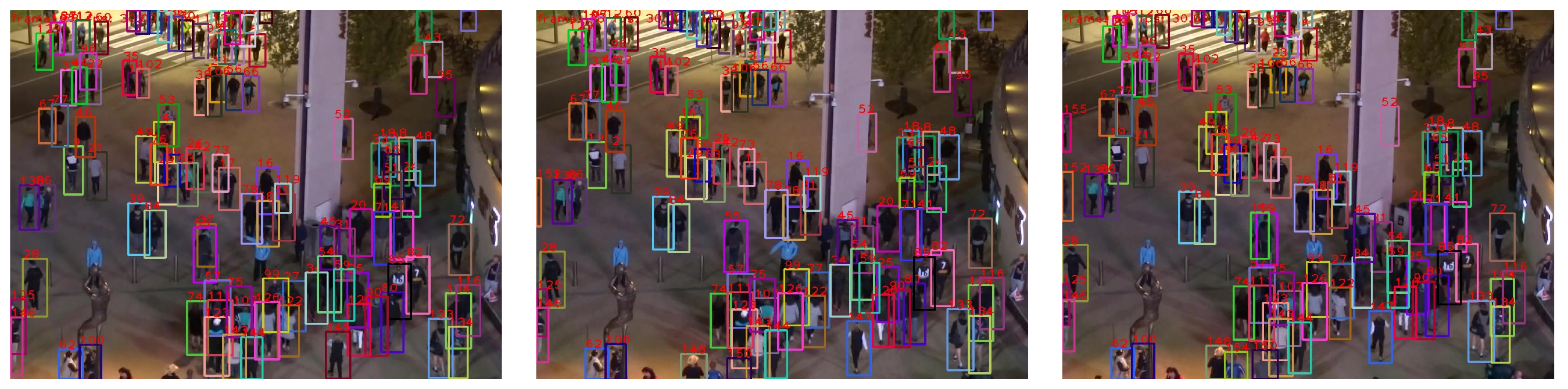}
            \label{fig5a}
	\end{minipage}
        }
        \\
        \subfloat[Non-linear motion Scene]{
	\begin{minipage}[t]{1\linewidth}
		  \includegraphics[width=0.98\columnwidth,keepaspectratio]{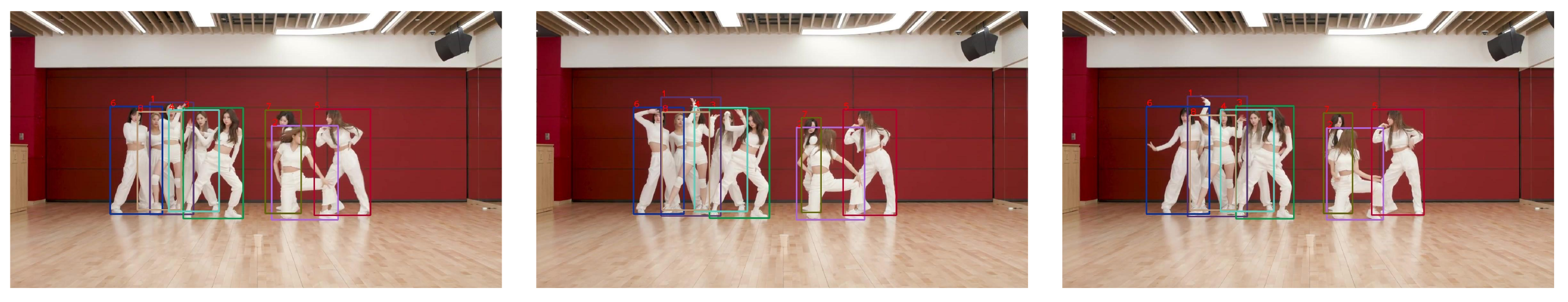}
            \label{fig5b}
		\end{minipage}
        }
        \caption{Tracking trajectories visualization of  and crowded scene in MOT20 and non-linear motion in Dancetrack.}
	\label{fig5}
        \vspace{-10pt}
\end{figure}

\section{Visualization}
\label{Visualization}
We offer visualizations of prototypical challenging scenes in MOT, including the non-linear scene (Figure \ref{fig5b}) and very crowded scene (Figure \ref{fig5a}), to demonstrate the tracking abilities of the proposed DiffusionTrack. We observe that our approach has a strong discriminative ability for object with severe non-linear motion and keeps high reliably associative ability in crowded scenes with dramatic occlusions.

\begin{algorithm}[htpb]
\SetAlgoLined
\DontPrintSemicolon 
\SetNoFillComment
\footnotesize
\KwIn{A video sequence $\texttt{V}$; diffusion track $\texttt{DT}$; association score threshold $\tau_{conf}$; detection score threshold $\tau_{det}$; number of boxes for association $N_{a}$;}
\KwOut{Tracks $\mathcal{T}_{activated}$ of the video}

Initialization: $\mathcal{T}_{activated},\mathcal{T}_{lost} \leftarrow \emptyset$\;
\For{frame $(f_{k-1},f_k)$ in $\texttt{V}$}{
	\tcc{predict association detection boxes \& scores}
	$\mathcal{D}_k \leftarrow \texttt{DT}(f_{k-1},f_k)$ \;
	$\mathcal{D}_{pre} \leftarrow \emptyset$ \;
	$\mathcal{D}_{cur} \leftarrow \emptyset$ \;
 	$\mathcal{D}_{new} \leftarrow \emptyset$ \;
	\For{$(idx,d_{k-1},d_k)$ in $\mathcal{D}_k$}{
	\If{$d_{k}.score > \tau_{conf}$}{
        \If{$idx < N_{a}$}{
	$\mathcal{D}_{pre} \leftarrow  \mathcal{D}_{pre} \cup \{d_{k-1}\}$ \;
    $\mathcal{D}_{cur} \leftarrow \mathcal{D}_{cur} \cup \{d_{k}\}$ \;
	}
        \ElseIf{$idx > N_{a}$}{
	$\mathcal{D}_{new} \leftarrow          \mathcal{D}_{new} \cup \{d_{k-1},d_{k}\}$ \;
	}
 }
	}

    \BlankLine
    \BlankLine

	\tcc{tracking association}
	Associate $\mathcal{T}_{activated}$ and $\mathcal{D}_{pre}$ using IoU
	Similarity\;
	$\mathcal{T}_{act-remain} \leftarrow \text{updating and remaining tracks from } \mathcal{D}_{cur}$ \;
 
    \BlankLine
	\BlankLine
 	\tcc{filter duplicated detection} 
	$\mathcal{D}_{new}  \leftarrow \mathcal{D}_{new} \setminus \mathcal{D}_{new}\cap \mathcal{D}_{cur}$ \;
    \BlankLine
	\BlankLine
 \tcc{predict new locations of lost tracks}
    \For{$t$ in $\mathcal{T}_{lost}$}{
	$t \leftarrow  KalmanFilter(t)$ \;
	}
     \BlankLine
    \BlankLine

	\tcc{tracking association}
	Associate $\mathcal{T}_{lost}$ and $\mathcal{D}_{new}$ using IoU
	Similarity\;
 	$\mathcal{D}_{remain} \leftarrow \text{remaining object boxes from } \mathcal{D}_{new}$ \;
	$\mathcal{T}_{lost-remain} \leftarrow \text{remaining tracks from } \mathcal{T}_{lost}$ \;
    \BlankLine
	\BlankLine
	\tcc{update activated and lost tracks}
	$\mathcal{T}_{activated} \leftarrow \mathcal{T}_{activated} \setminus \mathcal{T}_{act-remain} \cup \mathcal{T}_{lost} \setminus \mathcal{T}_{lost-remain}$ \;
 	$\mathcal{T}_{lost} \leftarrow \mathcal{T}_{act-remain} \cup  \mathcal{T}_{lost-remain}$ \;
	
    \BlankLine
	\BlankLine
	\tcc{initialize new tracks}
    \For{$d$ in $\mathcal{D}_{remain}$}{
	\If{$d.score > \epsilon$}{
	$\mathcal{T}_{activated} \leftarrow  \mathcal{T}_{activated} \cup \{d\}$ \;
	}
	}

        \BlankLine
	\BlankLine
	\tcc{reassociate lost tracks}
}
Return: $\mathcal{T}$
\caption{Pseudo-code of DiffusionTrack.}
\label{algorithm}
\end{algorithm}

\section{Inference Details}
\label{Inference Detail}
Once the association results are derived, IoU is utilized as the similarity metric to connect the object tracklets. To address potential occlusions, a simple Kalman filter is implemented to reassociate lost objects.
The pseudo-code of DiffusionTrack is shown in following Algorithm \ref{algorithm}.
\newline
\newline

\section*{Acknowledgments}
Min Yang was supported by National Key Research and Development Program of China (2022YFF0902100), Shenzhen Scienceand Technology Innovation Program (KOTD20190929172835662). Shenzhen Basic Research Foundation (JCYJ20210324115614039 and JCYJ20200109113441941). The computation is completed in the HPC Platform of Huazhong University of Science and Technology. 

\bibliography{aaai24}

\end{document}